\documentclass[conference]{IEEEtran}
\IEEEoverridecommandlockouts
\usepackage{cite}
\usepackage{amsmath,amssymb,amsfonts}
\usepackage{algorithmic}
\usepackage{graphicx}
\usepackage{textcomp}
\usepackage{xcolor}

\usepackage{hyperref}
\usepackage{multirow}
\usepackage{makecell}
\usepackage{graphicx}
\usepackage{comment}
\usepackage{xspace}
\usepackage{xcolor}
\usepackage{array}
\usepackage{caption}
\usepackage{subcaption}
\usepackage{wrapfig}
\usepackage{mathtools}
\usepackage{amsfonts}
\usepackage{caption,tabularx,booktabs}
\usepackage[final]{pdfpages}
\newcommand{\ourmethodFull}{\textsc{\textbf{D}ynamic \textbf{E}mbedding of Health\textbf{C}are \textbf{Ent}ities}\xspace}

\newcommand{\ourmethod}{\textsc{DECEnt}\xspace}
\newcommand{\ourmethodp}{\textsc{DECEnt+}\xspace}
\newcommand{\domain}{\textsc{Domain}\xspace}
\newcommand{\nodevec}{\textsc{Node2Vec}\xspace}
\newcommand{\deepwalk}{\textsc{DeepWalk}\xspace}
\newcommand{\rnn}{\textsc{RNN}\xspace}
\newcommand{\lstm}{\textsc{LSTM}\xspace}
\newcommand{\dynamictriad}{\textsc{CTDNE}\xspace}
\newcommand{\jodie}{\textsc{JODIE}\xspace}

\newcommand{\entities}{\mathcal{\tau}}
\newcommand{\doc}{\mathcal{D}}
\newcommand{\pat}{\mathcal{P}}
\newcommand{\room}{\mathcal{R}}
\newcommand{\med}{\mathcal{M}}

\newcommand{\Int}{\mathcal{S}}
\newcommand{\Med}{\mathcal{MD}}
\newcommand{\Pro}{\mathcal{PR}}
\newcommand{\Tra}{\mathcal{TR}}

\newcommand{\ps}{\mathbf{p}}
\newcommand{\pd}{\hat{\mathbf{p}}}
\newcommand{\ed}{\hat{\mathbf{e}}}
\newcommand{\pred}{\tilde{\mathbf{e}}}
\newcommand{\es}{\bar{\mathbf{e}}}

\newcommand{\W}{\mathbf{W}}

\newcommand{\prodmod}{\textsc{PM}\xspace}
\newcommand{\medmod}{\textsc{MM}\xspace}
\newcommand{\transmod}{\textsc{TM}\xspace}
\newcommand{\predmod}{\textsc{PRED}\xspace}

\newtheorem{problem}{Problem}

\def\BibTeX{{\rm B\kern-.05em{\sc i\kern-.025em b}\kern-.08em
    T\kern-.1667em\lower.7ex\hbox{E}\kern-.125emX}}

\begin{document}

\title{Dynamic Healthcare Embeddings for Improving Patient Care
}

\author{\IEEEauthorblockN{Hankyu Jang}
\IEEEauthorblockA{\textit{Department of Computer Science} \\
\textit{University of Iowa}\\
Iowa City, Iowa, USA \\
hankyu-jang@uiowa.edu}
\and
\IEEEauthorblockN{Sulyun Lee}
\IEEEauthorblockA{\textit{Interdisciplinary Graduate Program in Informatics} \\
\textit{University of Iowa}\\
Iowa City, Iowa, USA \\
sulyun-lee@uiowa.edu}
\and
\IEEEauthorblockN{D.~M.~Hasibul Hasan}
\IEEEauthorblockA{\textit{Department of Computer Science} \\
\textit{University of Iowa}\\
Iowa City, Iowa, USA \\
dmhasibul-hasan@uiowa.edu}
\and
\IEEEauthorblockN{Philip M.~Polgreen}
\IEEEauthorblockA{\textit{Department of Internal Medicine} \\
\textit{University of Iowa}\\
Iowa City, Iowa, USA \\
philip-polgreen@uiowa.edu}
\and
\IEEEauthorblockN{Sriram V.~Pemmaraju}
\IEEEauthorblockA{\textit{Department of Computer Science} \\
\textit{University of Iowa}\\
Iowa City, Iowa, USA \\
sriram-pemmaraju@uiowa.edu\\
*For the CDC MInD Healthcare Network
}
\and
\IEEEauthorblockN{Bijaya Adhikari}
\IEEEauthorblockA{\textit{Department of Computer Science} \\
\textit{University of Iowa}\\
Iowa City, Iowa, USA \\
bijaya-adhikari@uiowa.edu}
}

\maketitle

\begin{abstract}
As hospitals move towards automating and integrating their computing systems, more fine-grained hospital operations data are becoming available.
These data include hospital architectural drawings, logs of interactions between patients and healthcare professionals, prescription data, procedures data, and data on patient admission, discharge, and transfers.
This has opened up many fascinating avenues for healthcare-related prediction tasks for improving patient care. 
However, in order to leverage off-the-shelf machine learning software for these tasks, one needs to learn structured representations of entities involved from heterogeneous, dynamic data streams.
Here, we propose \ourmethod, 
an auto-encoding heterogeneous co-evolving dynamic neural network,
for learning heterogeneous dynamic embeddings of patients, doctors, rooms, and medications from diverse data streams.
These embeddings capture similarities among doctors, rooms, patients, and medications based on static attributes and dynamic interactions.

\ourmethod enables several applications in healthcare prediction, such as 
predicting mortality risk and case severity of patients,
adverse events (e.g., transfer back into an intensive care unit),
and future healthcare-associated infections.
The results of using the learned patient embeddings in predictive modeling show that \ourmethod has a gain of
up to 48.1\% on the mortality risk prediction task,
12.6\% on the case severity prediction task,
6.4\% on the medical intensive care unit transfer task, 
and 3.8\% on the Clostridioides difficile (C.diff) Infection (CDI) prediction task
over the state-of-the-art baselines.
In addition, case studies on the learned doctor, medication, and room embeddings show that our approach learns meaningful and interpretable embeddings.

\end{abstract}


\section{Introduction}
\label{section:introduction}

The availability of large-scale, high-resolution hospital operations data has opened up many avenues for the use of predictive modeling to improve patient care. Questions such as ``How likely is a patient at risk for an adverse event or complication that might require transfer into a Medical Intensive Care Unit (MICU) from another unit?'', ``Is a particular patient at risk of developing a \textit{healthcare-associated infection} (HAI) during their visit?'', ``Does a patient have a high risk of mortality?''. These are just a few examples of key predictive tasks which could help healthcare practitioners provide informed and personalized patient care. 

Recent advances in machine learning have enabled predictions in numerous domains. However, machine learning models such as recurrent neural networks, graph convolution networks, and transformers, often leveraged for predictive tasks, assume that the input data is structured. Unfortunately, hospital operations data consists of diverse data types, including architectural diagrams, admission-discharge-transfer logs, inpatient-doctor interactions, room visits, prescriptions, clinical notes, etc. These data tend to be unstructured and high-dimensional. Hence, learning structured representations of entities such as patients, doctors, rooms, and medications from hospital operations data is a key step in enabling the use of off-the-shelf machine learning algorithms for predictive tasks in healthcare settings.

\begin{figure*}
    \centering
    \includegraphics[width=\textwidth]{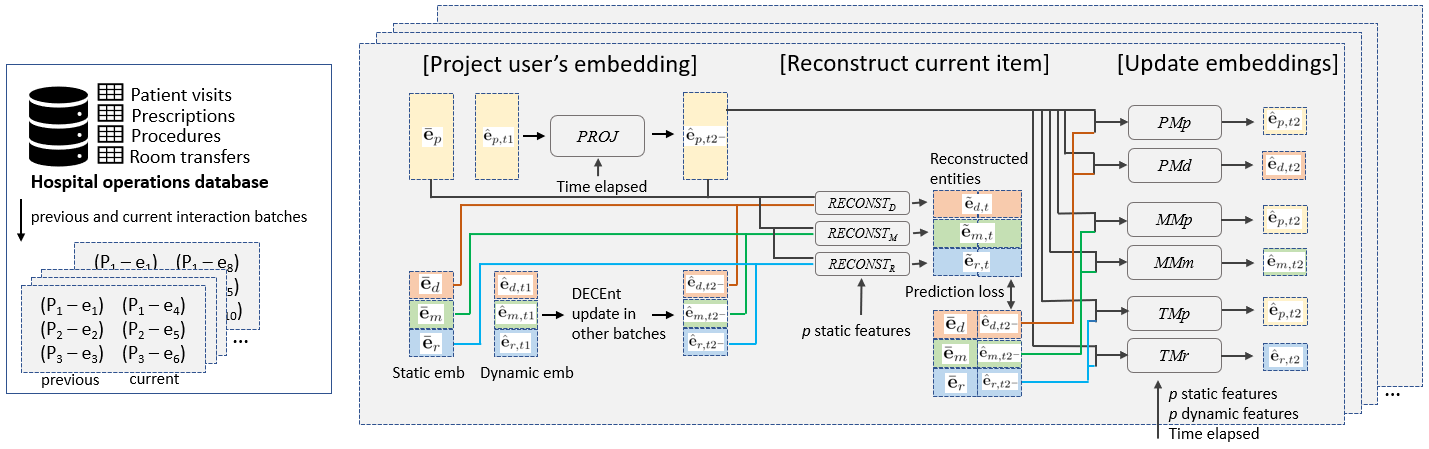}
    \caption{Our overall framework. 
    Our approach \ourmethod learns dynamic embeddings of healthcare entities from heterogeneous interaction log and other hospital operations data.
    For an interaction between a patient $p$ and an entity $e$ at time $t_2$,
    $p$'s dynamic embedding at time $t_1$ (time of $p$'s previous interaction) is projected to time $t_2^-$ (right before $t_2$).
    Then, the static and dynamic embeddings of $p$ and $e$ at time $t_2^-$ and $p$'s static features are used to reconstruct static and dynamic embeddings of entity $e$.
    Finally, dynamic embeddings of $p$ and $e$ get updated to time $t_2$ via update modules.
    We train \ourmethod in batches of interactions in parallel, while maintaining the temporal order of interactions across batches.
    }
    \label{fig:overall_framework}
\end{figure*}

There has been some recent interest in learning embeddings in a healthcare setting, primarily focusing on patient embeddings, such as MiME \cite{choi2018mime}. 
A major drawback of these approaches is that they all learn \textit{static} embeddings. In a healthcare setting, events that occur over time (e.g., prescription of an antibiotic, transfer to the MICU, and exposure to infection) play a crucial role, and a single static embedding is not representative enough. For example, a patient with a low risk of HAI at admission time could be considered high risk at a later day if there are patients with HAI in her unit. However, she could be considered low risk again in the future if she is treated with appropriate medications and shows no symptoms. Thus, learning a single embedding for the entire duration of the patient's stay does not capture the evolving nature of the risks involved and the care patient has received. 

Furthermore, none of the existing research on patient embeddings takes interactions between patients and other healthcare entities (e.g., physicians, hospital rooms, and medications) into account. However, capturing these interactions in learned embeddings is critical for numerous healthcare-associated predictive tasks. As an example, consider two interactions that happen in quick succession: (i) a patient $p_1$ in hospital room $r_1$ is prescribed the antibiotic vancomycin, which is commonly used as a treatment for suspected CDI\footnote{
CDI is shorthand for \emph{Clostridioides difficile infection}, a  healthcare-associated infection that affects gastro-intestinal regions leading to inflammation of the colon and severe diarrhea due to disruption of normal healthy bacteria in the colon \cite{CDCCDI}.}, (ii) a patient $p_2$ transfers into hospital room $r_2$ that is in the same unit as room $r_1$. Together, these interactions indicate an elevated risk of C.~diff infection for patient $p_2$, and we want the embeddings we learn to capture this.

There is also a separate thread of research on learning general-purpose dynamic embeddings (e.g., \cite{dai2016deep,kumar2019predicting}). 
However, these approaches learn embeddings from \textit{homogeneous} interactions between a user and a predefined item type. Such an approach is not readily applicable in a healthcare setting, where a patient may interact with heterogeneous entities, including physicians, medications, and rooms, which have different impacts on patients. Hence lumping all of these together into a single entity type will limit the discriminative power of the learned embeddings. Another line of related approaches includes dynamic network embeddings
\cite{zhou2018dynamic}.
However, these approaches are not readily applicable in our setting as they usually require a coarse snapshot representation of the network. 

To address the gap between existing approaches and the requirements of healthcare predictive modeling, we propose \ourmethod to learn dynamic embedding of entities associated with healthcare based on heterogeneous interactions.
\ourmethod learns general-purpose embedding in an unsupervised way which can be used in various predictive modeling tasks that can not be obtained via supervised training.
Specifically, \ourmethod jointly learns dynamic embeddings of patients, doctors, medications, and rooms while preserving hierarchical relationships between medications, doctor specializations, and physical proximity between rooms. 
In order to do so, \ourmethod maintains and updates weights for each specific interaction type and minimizes intra-entity similarity loss 
while performing an auto-encoder training scheme so that the embeddings of patient-entity interactions can re-construct their original features.
Our overall framework is presented in Figure \ref{fig:overall_framework}. Our contributions in this paper are as follows: 

\begin{itemize}
    \item We propose \ourmethod, a novel approach for learning dynamic embeddings of healthcare entities from heterogeneous interactions.
    \item \ourmethod enables several healthcare predictive modeling applications, including adverse event prediction, such as transfer to MICU, case severity and mortality prediction.
    \item We perform extensive experiments for evaluating patient embeddings. Our results show that \ourmethod outperforms state-of-the-art baselines in all the healthcare predictive modeling tasks we consider. Moreover, our embeddings are interpretable and meaningful.
\end{itemize}

\section{Preliminaries}
\label{section:preliminaries}
In this section, we give our setup, introduce the notations used throughout the paper, and finally state the problem formally.

\subsection{Setup and Notation}
\label{sec:setup}
Assume we are given a hospital operations database with a record of events on healthcare entities. The set of all healthcare entities $\entities$ includes the set of Doctors $\doc$, the set of Patients $\pat$, the set of medications $\med$, and the set of hospital rooms $\room$. 

The database has a record of time-stamped interactions between entities for each medical event that occurred in the hospital. In this paper, we consider three types of interactions. A \textbf{physician} interaction $(p,d,t)$, with $t \in \mathbb{R}^+$ and $0 \leq t \leq T$, represents that a doctor $d \in \doc$ performed a medical procedure on patient $p \in \pat$ at time $t$. Similarly, a \textbf{medication} 
interaction $(p,m,t)$ indicates that medicine $m \in \med$ was prescribed to patient $p \in \pat$ at time $t$, and finally a \textbf{spatial} interaction $(p,r,t)$ represents that a patient $p \in \pat$ was transferred to a hospital room $r \in \room$ at time $t$. We represent sets of physician, medication, and spatial interactions as $\Pro$, $\Med$, and $\Tra$, respectively. 

The database also consists of relationships among entities of the same type, represented as static graphs $G_{\mathrm{entitytype}}$. These are as follows:

\begin{itemize}
    \item \textbf{Room graph: } The architectural layout of the hospital can be represented as a graph $G_{\mathrm{room}}(\room,E_{\mathrm{room}})$ between hospital rooms.
    Two rooms $r_1$ and $r_2$ are connected by an edge $(r_1,r_2)$ in $G_{\mathrm{room}}$ if they are adjacent to each other.
    \item \textbf{Medication graph: }The medication hierarchy can be represented as a tree $G_{\mathrm{med}}(\med\cup{M}, E_{\mathrm{med}})$. Note that each leaf $m$ in the tree is a medication, i.e., $m \in \med$. The intermediate nodes $M$ represent medication sub-types.
    \item \textbf{Doctor graph: } Each doctor $d\in \doc$ has a specialty. We create a graph $G_{\mathrm{doc}}(\doc, E_{\mathrm{doc}})$, where the edges $(d_1,d_2)$ are based on the proximity of the specialty of doctors $d_1$ and $d_2$. 
\end{itemize}

Finally, for each patient $p \in \pat$, we are given static attributes $\ps_p$ with demographic and medical information. 
We are also given dynamic attributes $\pd_{p,t}$ at time $t$ that include information on the length of hospital stay, cumulative antibiotics count, gastric acid suppressors, and others.

In what follows, the lower case bold letters such as $\mathbf{v}$ represent vectors, the uppercase bold letters such as $\mathbf{W}$ represent matrices, and calligraphic symbols such as $\mathcal{S}$ represent sets. The lower case bold letters with hat such as $\hat{\mathbf{v}}$ represent dynamic vectors, and $\hat{\mathbf{v}}_t$ represents the vector at time $t$. Functions are represented by lowercase letters followed by braces, for example, $f(\cdot)$.

\subsection{Problem Statement}

Having defined the notations, we can now state our problem. 
Our goal is to learn dynamic embeddings for all patients, medications, rooms, and doctors 
based on the set of all interactions $\Int = \Pro \cup \Med \cup \Tra$. 
Without the loss of generality, we assume that the interactions in $\Int$ are sorted by time.  
We also aim to preserve the relationship among the entities of the same type represented by the static graphs in the embedding space.  Formally, the problem can be stated as follows:

\begin{problem} \textsc{Dynamic Healthcare Embeddings Problem}\\
\par\noindent\textbf{Given:} A set $\Int$ of time-stamped interactions among healthcare entities,  static networks $G_{\mathrm{room}}$, $G_{\mathrm{med}}$, $G_{\mathrm{doc}}$, and dynamic and static attributes of patients $\pd$ and $\ps$.
\par \noindent\textbf{Learn:} Dynamic embeddings 
$\ed_{u,t}$ for each entity $u \in \pat$ and
$\ed_{v,t}$ for each entity $v \in \doc \cup \med \cup \room$ and for each time $t$.
\par \noindent\textbf{Such that:} A function $f(\ed_{u,t}, \ed_{v,t})$ encodes information to be predictive of $v$,
and the $dist(\ed_{v,t}, \ed_{v',t})$  between the embeddings $\ed_{v,t}$ and $\ed_{v',t}$ is of two entities of the same type $v$ and $v'$ reflects the distance between the two in $G_{\mathrm{entity type}}$.
\label{prob:problem}
\end{problem}

\section{Method}
\label{section:method}
Here we propose our method $\ourmethod$ (short for \ourmethodFull), a novel auto-encoding heterogeneous co-evolving dynamic neural network, to solve Problem \ref{prob:problem}. \ourmethod models heterogeneous interactions using a set of jointly learned co-evolving networks.
It has three main components at a high level, each of which models one of the interaction types.
As the interactions occur, the dynamic embeddings of the entities involved are updated simultaneously. 
These dynamic embeddings are jointly trained to reconstruct the entity that the patients interacts with while preserving similarities imposed by $G_{\mathrm{entity type}}$.
We describe each of the components in detail next.

\subsection{Physician Module}

Consider a physician interaction $(p,d,t) \in \Pro$ between a patient $p$ and a doctor $d$ at time $t$. To reflect that $p$ and $d$ interacted with each other, our method simultaneously updates the maintained dynamic embeddings $\ed_{p,t}$ and $\ed_{d,t}$ of both $p$ and $d$.
To do so, we design a pair of co-evolving deep neural networks $\prodmod_p$ and $\prodmod_{d}$ to update the patient's and doctor's embedding, respectively.  

Following the literature in the co-evolutionary neural networks~\cite{kumar2019predicting, dai2016deep}, we design $\prodmod_p$ and $\prodmod_{d}$ to be mutually recursive, i.e., the patient's embedding at time $t$, $\ed_{p,t}$ generated by the  $\prodmod_p$ depends on both the patient's embedding at time $t^-$ (just before time $t$) $\ed_{p,t^-}$ and the doctor's embedding $\ed_{d,t^-}$ prior to the interaction as well as the time elapsed between the patient's previous and the current interaction $\Delta_{p,t}$. 

However, relying only on interactions is not enough for patient care predictive modeling applications. While the interactions capture the medical events of a particular patient, they do not capture the inherent risk the patient is at due to the patient's own underlying health conditions and demographic features. Hence, we ensure that the patient's dynamic embedding $\ed_{p,t}$ also depends on the patient's static feature $\ps_p$ and the dynamic feature $\pd_{p,t}$ at time $t$. 
We ensure that the doctor's dynamic embedding $\ed_{d,t}$ depends on $\ps_p$ and $\pd_{p,t}$ to better track all the patients the doctor $d$ has interacted with.

Following are the update equations for  $\prodmod_p$ and  $\prodmod_d$ respectively.



\begin{equation}
\small
\label{eq:hidep_update}
\begin{aligned}
\ed_{p,t} &=\sigma\left[ \W_{p}^{PM} [\ed_{p,t^-} \mathbin| \ed_{d,t^-} \mathbin| \Delta_{p,t} \mathbin| \ps_p  \mathbin| \pd_{p,t}] + \mathbf{B}_{p}^{PM} \right] \\
\ed_{d,t} &=\sigma\left[ \W_{d}^{PM} [\ed_{d,t^-} \mathbin| \ed_{p,t^-} \mathbin| \Delta_{d,t} \mathbin| \ps_p  \mathbin| \pd_{p,t}] + \mathbf{B}_{d}^{PM} \right]
\end{aligned}
\end{equation}
Here, $\W_{p}^{PM}$ is the weight matrix parameterizing $\prodmod_p$ and
$\W_{d}^{PM}$ is the weight matrix for $\prodmod_d$.
$\mathbf{B}_{p}^{PM}$ and $\mathbf{B}_{d}^{PM}$ are bias. 
$[\mathbf{a} \mathbin| \mathbf{b}]$ denotes concatenation of the two vectors $\mathbf{a}$ and $\mathbf{b}$. Finally $\sigma$ is a non-linear activation function, and in our experiments, we set $\sigma$ to be the tanh activation.

Note that $\ed_{p,t^-}$ in Equation \ref{eq:hidep_update} is the patient embedding immediately prior to the interaction $(p,d,t)$. In earlier related approaches,  $\ed_{p,t^-}$ is taken to be the embedding generated by $\prodmod_p$ for patient $p$'s last interaction before time $t$. However, in cases where the gap between consecutive interactions are too large, this approach is sub-optimal. Hence, in this paper, we use the embedding projection operation~\cite{kumar2019predicting}. Specifically, a patient $p$'s embedding after time $\Delta$ is projected to be,
\begin{equation}
    \ed_{p,t + \Delta} = (1 + \W\times \Delta) + \ed_{p,t}
    \label{eq:projection}
\end{equation}
where $\W$ is a linear weight matrix. We use Equation \ref{eq:projection} to generate $\ed_{p,t^-}$ by projecting the embedding generated by $\prodmod_p$ for $p$'s previous interaction.

\subsection{Medication Module}
Similar to the Physician module, we maintain a pair of co-evolving deep neural networks $\medmod_{p}$ and $\medmod_{m}$ to update the embeddings $\ed_{p,t}$ and $\ed_{m,t}$ post a medication interaction $(p,m,t)$. Recall that a medication interaction $(p,m,t) \in \Med$ represents that a medication $m$ was prescribed to a patient $p$ at time $t$. Here, $\medmod_{p}$ updates the dynamic embedding of $p$ at time $t$ and $\medmod_{m}$ does the same for medicine $m$. The update equation for the medication module are similar to the physician module,

\begin{equation}
\small
\label{eq:medication_update}
\begin{aligned}
\ed_{p,t} &=\sigma\left[ \W_{p}^{MM} [ \ed_{p,t^-} \mathbin| \ed_{m,t^-} \mathbin| \Delta_{p,t} \mathbin| \ps_p  \mathbin| \pd_{p,t} ] + \mathbf{B}_{p}^{MM}  \right] \\
\ed_{m,t} &=\sigma\left[ \W_{m}^{MM} [\ed_{m,t^-} \mathbin| \ed_{p,t^-} \mathbin| \Delta_{m,t} \mathbin| \ps_p \mathbin| \pd_{p,t} ] + \mathbf{B}_{m}^{MM} \right]
\end{aligned}
\end{equation}
where $\W_{p}^{MM}$ and $\W_{m}^{MM}$ are the weight matrices for $\medmod_p$ and $\medmod_m$, respectively. $\mathbf{B}_{p}^{MM}$ and $\mathbf{B}_{m}^{MM}$ are bias.

\subsection{Transfer Module}
Similar to the previous two modules, the transfer module also consist of two co-evolving neural networks. 
$\transmod_{p}$ updates the patient embedding $\ed_{p,t}$  and $\transmod_{r}$ updates the room embedding post a transfer interaction $(p,r,t) \in \Tra$. The update equations for the transfer module are as follows:

\begin{equation}
\small
\label{eq:room_update}
\begin{aligned}
\ed_{p,t} &=\sigma\left[ \W_{p}^{TM} [\ed_{p,t^-} \mathbin| \ed_{r,t^-} \mathbin| \Delta_{p,t} \mathbin| \ps_p  \mathbin| \pd_{p,t} ] + \mathbf{B}_{p}^{TM}  \right] \\
\ed_{r,t} &=\sigma\left[ \W_{r}^{TM} [\ed_{r,t^-} \mathbin| \ed_{p,t^-} \mathbin| \Delta_{r,t} \mathbin| \ps_p  \mathbin| \pd_{p,t} ] + \mathbf{B}_{r}^{TM} \right]
\end{aligned}
\end{equation}
where $\W_{p}^{TM}$ and $\W_{r}^{TM}$ are the weight matrices for $\transmod_p$ and $\transmod_r$, respectively. $\mathbf{B}_{p}^{TM}$ and $\mathbf{B}_{r}^{TM}$ are bias.

\subsection{Reconstruction Module}
\label{sec:prediction}

Consider the interaction $(p,e,t_2)$ such that a patient $p$ interacted with an entity $e$ at time $t_2$ and that $p$'s previous interaction occurred at time $t_1$ such that $t_1 < t_2$.
To learn a meaningful representations of patient $p$ and an entity $e$ in the latent space, we train $\ourmethod$ to update embeddings of both $p$ and $e$ such that they are capable of reconstructing the embedding of $e$ at time $t_2$.
Specifically, while processing the interaction $(p,e,t_2)$, 
our goal is to learn dynamic embeddings $\ed_{p,t_2^-}$ and $\ed_{e,t_2^-}$ immediately before time $t_2$ such that they are useful in reconstructing both
dynamic embedding $\ed_{e,t_2^-}$  and the 
static embedding $\es_{e}$ of entity $e$.

In order to reconstruct the dynamic and the static embeddings of entity $e$, we feed in 
static embeddings $\es_{p}$ and $\es_{e}$ of patient $p$ and entity $e$ along with the projection of the generated 
dynamic embeddings $\ed_{p,t_2^-}$ and 
dynamic embeddings of $e$ that is updated via consecutive interactions with other patients $\ed_{e,t_2^-}$ to the reconstruction module.
Each entity type (e.g., doctor, medication, room) has its own reconstruction module (e.g., \textit{RECONST}\textsubscript{D},
\textit{RECONST}\textsubscript{M},
\textit{RECONST}\textsubscript{R}), 
which is defined as follows:
\begin{equation}
\small
\label{eq:hidep_predict}
\begin{aligned}
\pred_{d,t_2^-}&=\W_d \left[\ed_{p,t_2^-} \mathbin| \es_{p} \mathbin| \ps_{p} \mathbin| \ed_{d,t_2^-} \mathbin| \es_{d}\right] + \mathbf{B}_d\\
\pred_{m,t_2^-}&=\W_m \left[\ed_{p,t_2^-} \mathbin| \es_{p} \mathbin| \ps_{p} \mathbin| \ed_{m,t_2^-} \mathbin| \es_{m}\right] + \mathbf{B}_m\\
\pred_{r,t_2^-}&=\W_r \left[\ed_{p,t_2^-} \mathbin| \es_{p} \mathbin| \ps_{p} \mathbin| \ed_{r,t_2^-} \mathbin| \es_{r}\right] + \mathbf{B}_r\\
\end{aligned}
\end{equation}
where $\W_d$, $\W_m$, and $\W_r$ are the learnable weight matrices for reconstructing embeddings of doctor, medication, and room and $\mathbf{B}_d$, $\mathbf{B}_m$, and $\mathbf{B}_r$ are bias. Note that $\pred_{e,t_2^-}$ is the predicted embedding of size $|\ed_{e,t_2^-}| + |\es_{e}|$.

\ourmethod uses onehot vector for static embeddings.
We have another model \ourmethodp, which we use Bourgain embeddings \cite{Bourgain1986TheMI} for static embeddings of $v \in \doc \cup \med \cup \room$, which we compute from static graphs $G_{\mathrm{entity type}}$.

\subsection{Overall Framework}

While our primary goal is to train $\ourmethod$ to 
encode the entity that a patient interacts with,
we also want to enforce additional losses to ensure that the generated embeddings are interpretable. We analyze our learned embeddings with the help of a domain expert (See Section \ref{section:experiment}). We describe the losses used to train \ourmethod next.

\par\noindent\textbf{Reconstruction Loss:} \emph{Reconstruction loss} is encoded as the error between the predicted and the ground truth embeddings of the entity a patient interacts with.
Continuing with our example from Section \ref{sec:prediction}, we want to minimize the difference between 
the reconstructed embedding $\pred_{e,t_2^-}$ and the 
ground truth embedding $[\ed_{e,t_2^-} \mathbin| \es_{e_2}]$, where $\mathbin|$ is the concatenation operation. We enforce the reconstruction loss on all interactions including procedure interactions $\Pro$, medication interactions $\Med$, and transfer interactions $\Tra$.

Formally, reconstruction loss is defined as follows:
\begin{align}
 \small
    L_{reconst} &= \sum_{(p,d,t) \in \Pro} ||\pred_{d,t^-}  - [\ed_{d,t^-}|\es_{d}]||_2  \nonumber\\
    &+ \sum_{(p,m,t) \in \Med} ||\pred_{m,t^-}  - [\ed_{m,t^-}|\es_{m}]||_2   \nonumber \\
     &+ \sum_{(p,r,t) \in \Tra} ||\pred_{r,t^-}  - [\ed_{r,t^-}|\es_{r}]||_2
\end{align}
where $||\mathbf{v}||_2$ is the $L_2$ norm of the vector $\mathbf{v}$. 

\par\noindent\textbf{Temporal Consistency Loss:} We want to ensure that the embeddings of entities do not vary dramatically between consecutive interactions. To this end, we define \emph{temporal consistency} loss as the $L_2$ norm of the difference between the embeddings of each entity between each consecutive interaction.  Formally, it is defined as follows:

\begin{equation}
\small
    L_{temp} = \sum_{(p,e,t) \in \Int} ||\ed_{p,t} -\ed_{p,t^-}||_2 +  ||\ed_{e,t} -\ed_{e,t^-}||_2
\end{equation}

As defined earlier, $\Int$ is the set of all interactions, i.e., $\Int = \Pro \cup \Med \cup \Tra$. 

\par\noindent\textbf{Domain Specific Loss:}
Furthermore, we also implement a set of losses to ensure that the entities known to be similar as per domain knowledge have similar embeddings. To this end, we first compute the Laplacian matrices, $\mathbf{L}_{\mathrm{room}}$, $\mathbf{L}_{\mathrm{med}}$, and $\mathbf{L}_{\mathrm{doc}}$ corresponding to the static graphs, $G_{\mathrm{room}}$, $G_{\mathrm{med}}$, and $G_{\mathrm{doc}}$ representing similarities between the entities (see Section \ref{sec:setup} for details on the graphs). We then compute the graph Laplacian based \emph{domain specific loss} as follows:

\begin{align}
\small
    L_{dom} &= \lambda_{dom}^D \sum_{t \in [0,T], d \in \doc} \ed_{d,t}^T  \mathbf{L}_{\mathrm{doc}} \ed_{d,t}\\
    &+ \lambda_{dom}^M \sum_{t \in [0,T], m \in \med} \ed_{m,t}^T  \mathbf{L}_{\mathrm{med}} \ed_{m,t} \nonumber \\ 
    &+ \lambda_{dom}^R \sum_{t \in [0,T], r \in \room} \ed_{r,t}^T  \mathbf{L}_{\mathrm{room}} \ed_{r,t} \nonumber
\end{align}

Note that $\ed_{r,t}^T$ is a transpose of $\ed_{r,t}$ and $\lambda_{dom}^D$, $\lambda_{dom}^M$ and $\lambda_{dom}^R$ are scaling constants in the equation above. 
Note that the $L_{dom}$ decreases in magnitude if entities connected by an edge have similar embeddings.

\par \noindent \textbf{Overall Loss and Training: } Our overall loss is the weighted aggregation of the previous three losses. 
Note that we jointly train all three modules along with the reconstruction module after pre-training each individual module. 
We optimize the overall loss using the Adam optimization algorithm~\cite{kingma2014adam}.  
We use Adam optimizer with the learning rate of 1e-3 and the weight decay of 1e-5.
The size of the dynamic embeddings of \ourmethod and \ourmethodp are set to 128.
We train \ourmethod and \ourmethodp for 1000 epochs with an early stopping on the training loss with the patience of 10 epochs.


\section{Experiment}
\label{section:experiment}
\begin{figure*}[t!]
    \centering
    \begin{subfigure}[b]{0.31\textwidth}
         \centering
         \includegraphics[width=\textwidth]{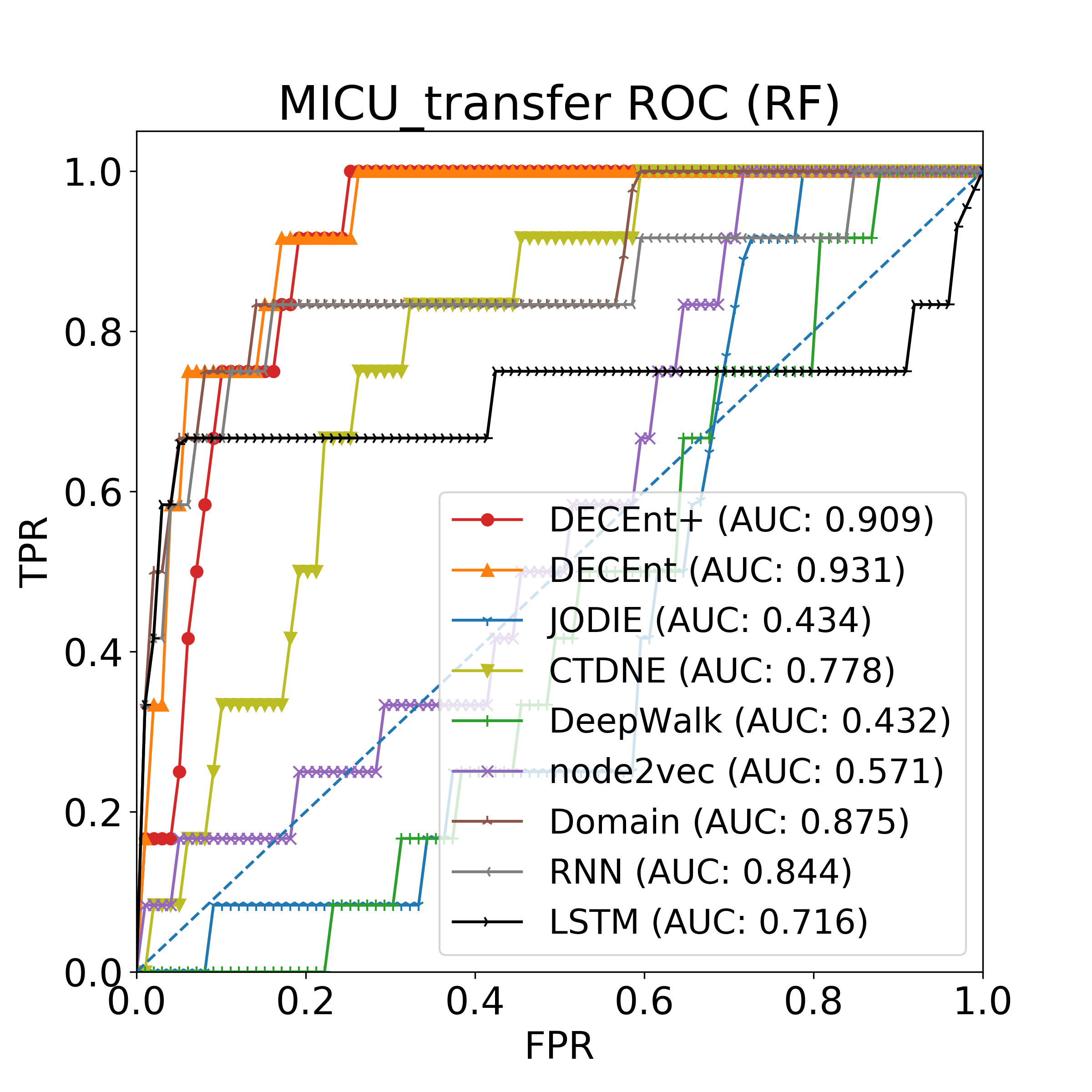}
         \label{fig:MICU_transfer_ROC_RF}
     \end{subfigure}
     \hfill
     \begin{subfigure}[b]{0.31\textwidth}
         \centering
         \includegraphics[width=\textwidth]{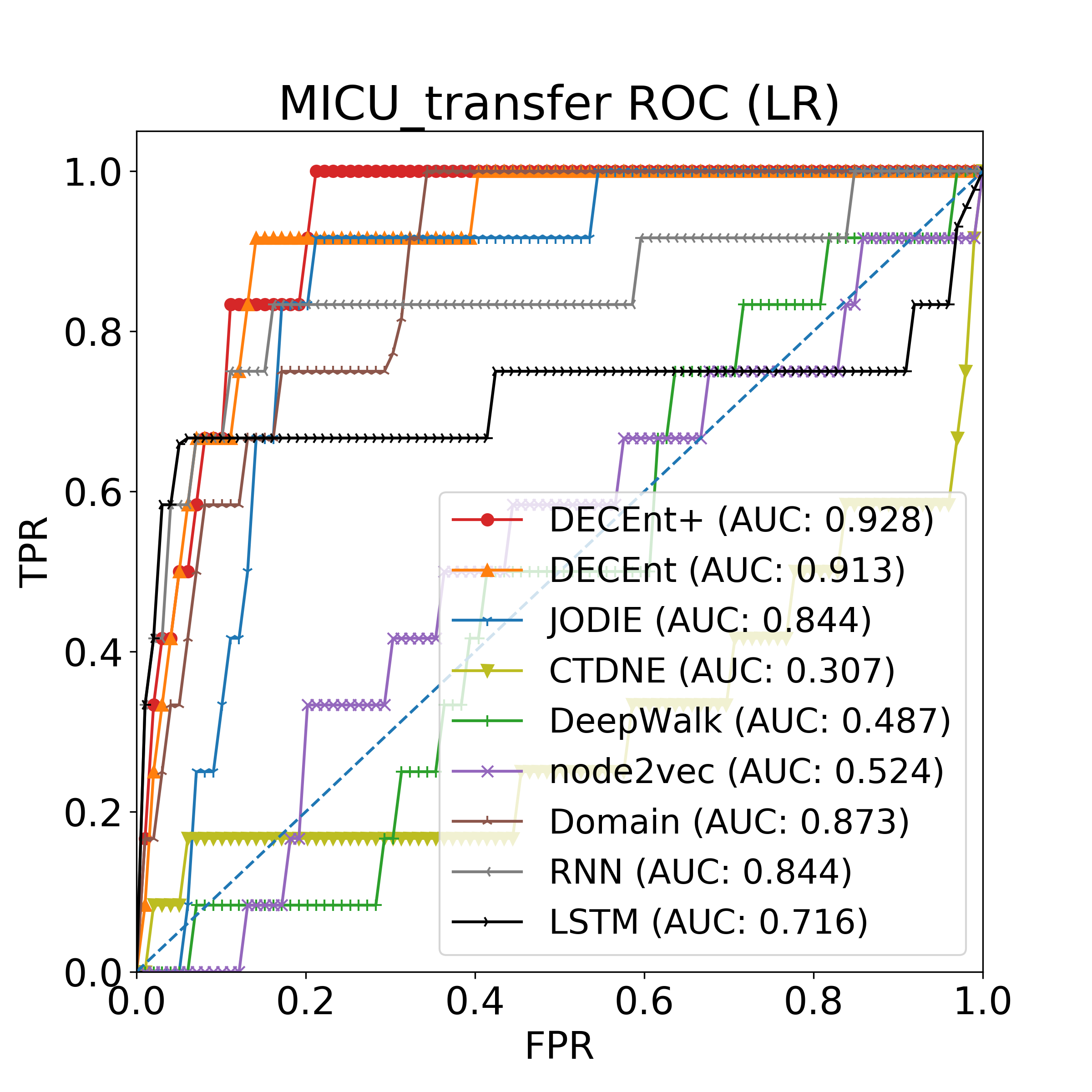}
         \label{fig:MICU_transfer_ROC_LR}
     \end{subfigure}
     \hfill
     \begin{subfigure}[b]{0.31\textwidth}
         \centering
         \includegraphics[width=\textwidth]{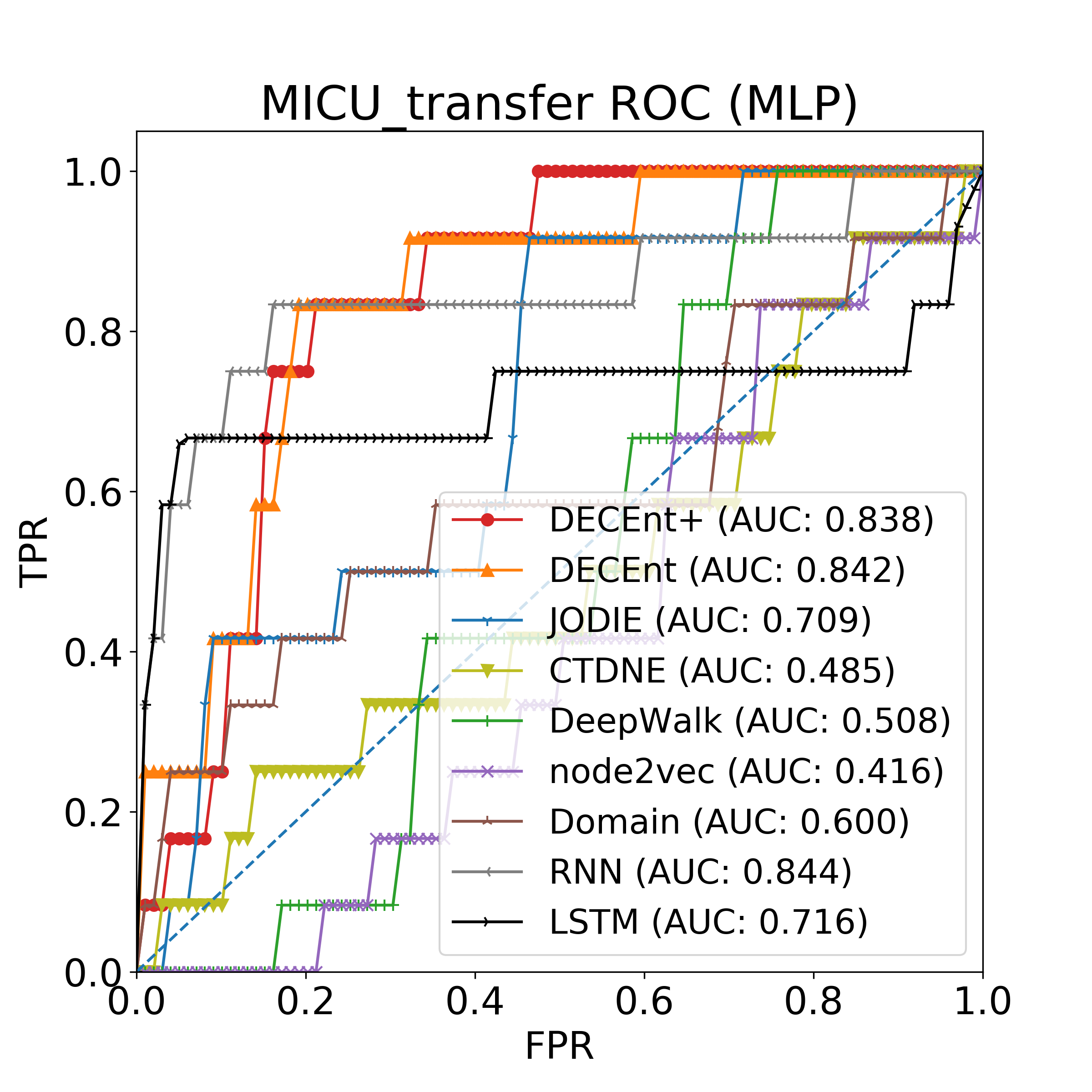}
         \label{fig:MICU_transfer_ROC_MLP}
     \end{subfigure}
\caption{ ROC curves for \ourmethod (in orange), \ourmethodp (in red) and the baselines 
for MICU transfer prediction task.
Each figure corresponds to a different classifier: random forest (left), logistic regression (middle) and multi-layer preceptron (right).
Both variants of \ourmethod outperform all other methods consistently.
}
\label{fig:MICU_transfer_ROC}
\end{figure*}

We describe our experimental setup next.
We provide code for academic purposes
\footnote{\url{https://github.com/HankyuJang/DECEnt-dynamic-healthcare-embeddings}}
Experiments were conducted on Intel(R) Xeon(R) machine with 528GB memory and 4 GPUs (GeForce GTX 1080 Ti).

\par \noindent \textbf{Data: }We evaluate \ourmethod on the real world hospitals operation data collected from University of Iowa Hospitals and Clinics (UIHC). UIHC is a large (800-bed) tertiary care teaching hospital located in Iowa City, Iowa.
The dataset consists of de-identified electronic medical records (EMR), and admission-discharge-transfer (ADT) records on
6,496 patients between January 01, 2010 and March 31, 2010.
Each patient visit has a set of diagnoses, a timestamped record on a set of medications prescribed, and a procedures performed by physicians. For each patient room transfer event during the visit, the source room and the destination room are recorded with a timestamp.
The 
patients in our dataset interacted with 575 doctors where 23,085 physician interactions were observed during the time-frame.
686 unique medicines were prescribed to patients, with a total of 349,345 medication interactions.
Moreover, patients visited 557 rooms, with a total of 16,771 spatial interactions.

\par \noindent \textbf{Baselines: }We compare performance of \ourmethod and \ourmethodp against natural and state-of-the-art baselines in all of our applications.
The first group of baselines include popular static network embeddings \nodevec~\cite{grover2016node2vec} and \deepwalk~\cite{perozzi2014deepwalk} and dynamic network embedding \dynamictriad~\cite{Nguyen2018CTDNE}.
After extensive literature review~\cite{li2019using, min2019predictive}, we find that predictive modelling tasks in healthcare analytics consist of an off-the-shelf classifier and a set of handcrafted feature. 
To this end, we design \domain baselines by augmenting classifiers with feature selection targeted for each predictive task. We also compare against
time-nested deep recurrent neural models \lstm and \rnn.
Our final baseline is the state-of-the-art co-evolutionary neural network \jodie \cite{kumar2019predicting}.


\subsection{Application 1: MICU Transfer Prediction}

The first application we consider seeks to forecast whether a patient is at risk of transfer to a Medical Intensive Care Unit (MICU). The MICU provides care for patients at a critical stage. A patient is only transferred to the MICU when there is a necessity for constant monitoring and intensive care. Therefore, an early indication of the high risk of transfer to a MICU can guide healthcare professionals to increase patient care. On the other hand, MICU beds are a scarce resource, and hence, an early indication of potential MICU transfer helps hospital officials allocate resources better.

Here we pose MICU transfer prediction as a binary classification problem.
The input to a classifier is the embedding generated by \ourmethod at time $t$.
The output is a label representing whether a patient will be transferred to a MICU at time $t+1$.
Here we only consider the cases where a Non-MICU to MICU transfer occurs at least three days from admission.
We construct the positive instances (+) based on the actual MICU transfer events.
If there is more than one transfer for a patient, we only consider the latest event.
We randomly sample one day of their embedding for patients with no such event and use it as negative instances (-).
 Note that the MICU transfers are rare events. In order to remain faithful to the problem, we ensure that there is a significant class imbalance of greater than 100:1.

We train three independent off-the-shelf classifiers, logistic regression (LR), random forest (RF), and multi-layer perceptrons (MLP), to predict MICU transfers.
We perform 5-fold cross-validation. Due to the extreme class imbalance, we randomly undersample the majority class training data to match the size of the minority class. However, we retain the class imbalance in the test set.
We use the area under the ROC curve, AUC, as an evaluation metric robust under class imbalance.
Finally, we report the average AUC over 30 repetitions and the standard deviation.
We repeat the entire process with the baseline methods. 
Results are presented in Fig \ref{fig:MICU_transfer_ROC}.

As seen in the 
figure, \ourmethod outperforms all the baselines regardless of the underlying classifier. The gain of \ourmethod over the most competitive baseline \domain is 6.4\%. 
A significant advantage \ourmethod has over the baselines, is that it is able to distinguish entity types from each other and it is specifically designed to model the heterogeneous dynamic interactions. 
The results show that \ourmethod indeed learns embeddings that are useful for predictive modelling. 
This highlights the importance of principally encoding the heterogeneous nature of interactions that occur in a healthcare setting. 

Another interesting observation we make is that \ourmethod outperforms \ourmethodp in two out of three classifiers implying that restrictively enforcing the embeddings to respect the domain specific distances may lead to poorer performance in some cases while being useful in other. Impressively, even in the classifier where \ourmethod performs the poorest, it still outperforms the most competitive baseline.


\subsection{Application 2: CDI Prediction}

CDI spreads in a healthcare setting, infecting patients who are already in a weakened state. Acquiring CDI increases a patient's mortality risk and prolongs hospital stay. Hence, early identification of patients at risk of CDI gives healthcare providers valuable lead time to combat the infection. 
It also helps prevent infection spread as practitioners can employ contact precaution and additional sanitary measures with special attention to the area around a patient at risk. Note that according to current practice, a patient is tested for CDI after three days of symptoms~\cite{monsalve2015improving}. This means that pathogen may have spread even before positive test results. 

Here we pose CDI prediction as a classification problem. 
The goal here is to accurately predict a label indicating if a patient would get infected within the next three days, given the embeddings learned by \ourmethod and baselines.
For the patients who get CDI, we use their embeddings three days before the positive report as positive class instances.
We randomly sample one day of their visit for each of the rest of the patients and use it as negative class instances.
We have a class imbalance of 89:1 as CDI cases are rare.
Our experimental setup is the same as Application 1. 
We present our results in Table \ref{tab:CDI}.

\begin{table}[h]
\centering
\caption{Average AUC and the corresponding s.t.dev. for various approaches for the CDI prediction task.
\ourmethod and \ourmethodp outperform all the baselines.
}
\label{tab:CDI}
\begin{tabular}{|c|c|c|c|}
\hline
Method & \multicolumn{3}{c|}{AUC} \\
\hline
\rnn & \multicolumn{3}{c|}{0.56 (0.119)} \\
\lstm & \multicolumn{3}{c|}{0.585 (0.103)} \\
\hline
-             & LR            & RF            & MLP           \\
\hline
\domain                     & 0.655 (0.123) & 0.709 (0.104) & 0.582 (0.137) \\
\deepwalk      & 0.494 (0.087) & 0.487 (0.093) & 0.492 (0.103) \\
\nodevec & 0.453 (0.098) & 0.43 (0.106)  & 0.478 (0.1)   \\
\dynamictriad & 0.463 (0.101) & 0.528 (0.079) & 0.483 (0.116)\\
\jodie                      & 0.552 (0.192) & 0.377 (0.177) & 0.469 (0.176) \\
\hline
\ourmethod       & 0.732 (0.069) & 0.711 (0.08)  & 0.668 (0.082) \\
\ourmethod+ & \textbf{0.736 (0.064)} & 0.717 (0.078) & 0.664 (0.091) \\
\hline
\multicolumn{4}{l}{$^{\mathrm{a}}$The value in bold denotes best performance}
\end{tabular}
\end{table}

As observed in the table \ourmethod and \ourmethodp outperform all the baselines with the gain over the best performing baseline \domain of 3.81\%. As in the previous applications static embedding approaches perform the worst, followed by the dynamic embedding baseline \dynamictriad, and finally \jodie.

\subsection{Application 3: Mortality and Case Severity Risk Prediction}

The Agency for Health Research and Quality (AHRQ) is a federal agency that performs mortality and case severity analysis on inpatient visits across hospitals in the US. Hospitals submit records on each patient visit to the agency. The agency then reports the severity and mortality along with the ``expected'' severity and mortality risk for each patient. 
The agency's report acts as a quality control metric for hospitals via retrospective analysis.
Predicting case severity and mortality while the patient is in the hospital has many applications, ranging from personalized patient care to resource allocation.

\begin{table}[h]
\centering
\caption{Average F1 Macro and the corresponding s.t.dev.
for various methods 
for mortality and severity prediction.
\ourmethod and \ourmethodp outperform all the baseline methods.}
\label{tab:mortality_severity_f1macro}
\begin{tabular}{|c|c|c|}
\hline
Method  & \textbf{Mortality}     & \textbf{Severity}      \\
\hline
\rnn     & 0.276 (0.039) & 0.31 (0.032)  \\
\lstm    & 0.289 (0.033) & 0.308 (0.026) \\
\domain  & 0.22 (0.017)  & 0.258 (0.007) \\
\deepwalk      & 0.172 (0.034) & 0.192 (0.019) \\
\nodevec & 0.172 (0.02)  & 0.196 (0.009) \\
\dynamictriad & 0.184 (0.019) & 0.199 (0.007) \\
\jodie   & 0.143 (0.039) & 0.193 (0.014) \\
\hline
\ourmethod  & 0.421 (0.027) & 0.34 (0.014)  \\
\ourmethodp & \textbf{0.428 (0.022)} & \textbf{0.349 (0.015)} \\
\hline
\multicolumn{3}{l}{$^{\mathrm{a}}$The value in bold denotes best performance}
\end{tabular}
\end{table}

AHRQ classifies each case into four categories, namely \emph{Minor}, \emph{Moderate}, \emph{Major}, and \emph{Extreme} for both mortality and case severity. 
Hence, we model both these tasks as multi-label classification problem.
The input to a classifier is the embedding generated by DECEnt by the time of discharge.
We report the results on logistic regression classifier on F1-Macro scores in Table \ref{tab:mortality_severity_f1macro}.
The results show that \ourmethod and \ourmethodp consistently outperform all the baselines for both mortality and case severity prediction, with a gain of up to 48.1\% over LSTM and 12.58\% over RNN, respectively, compared to the best performing baselines for each prediction task.
This result reinforces our conclusion from previous applications that \ourmethod is superior to state-of-the-art baselines in predictive modeling tasks in the healthcare setting.
\subsection{Evaluating the Embeddings}
Results described earlier in the paper show that embeddings obtained via \ourmethod are excellent for a variety of predictive tasks. 
In this subsection, we present findings from a qualitative exploration of the embeddings. 
This was done in collaboration with a medical doctor at the hospital who specializes in infectious diseases.

We start our exploration by computing \textit{dispersions} for subsets of healthcare entities.
Suppose that the set of doctors $\doc$ is partitioned into subsets $\doc_1, \doc_2, \ldots, \doc_d$.
For any $1 \le i < j \le d$ and time $t$ we define the pairwise \textit{doctor dispersion} between $\doc_i$ and $\doc_j$ as
\begin{align}
\small
    disp_{\doc, t}(i, j) = \frac{\sum_{d \in \doc_i, d' \in \doc_j} ||\ed_{d,t} - \ed_{d',t}||_2}{|\doc_i|\cdot |\doc_j|}
\end{align}

Recall that $\ed_{d,t}$ denotes the embedding of a doctor $d$ at time $t$.
If the partition $\doc_1, \doc_2, \ldots, \doc_d$ represents a grouping of doctors by specialty (e.g., pediatrics, cardiology, anesthesia, neurology, etc.) then the pairwise
doctor dispersions provide a measure of the average distance between doctors from different specialties in the computed embeddings.

\begin{figure}[t!]
\centering
\includegraphics[width=0.8\columnwidth]{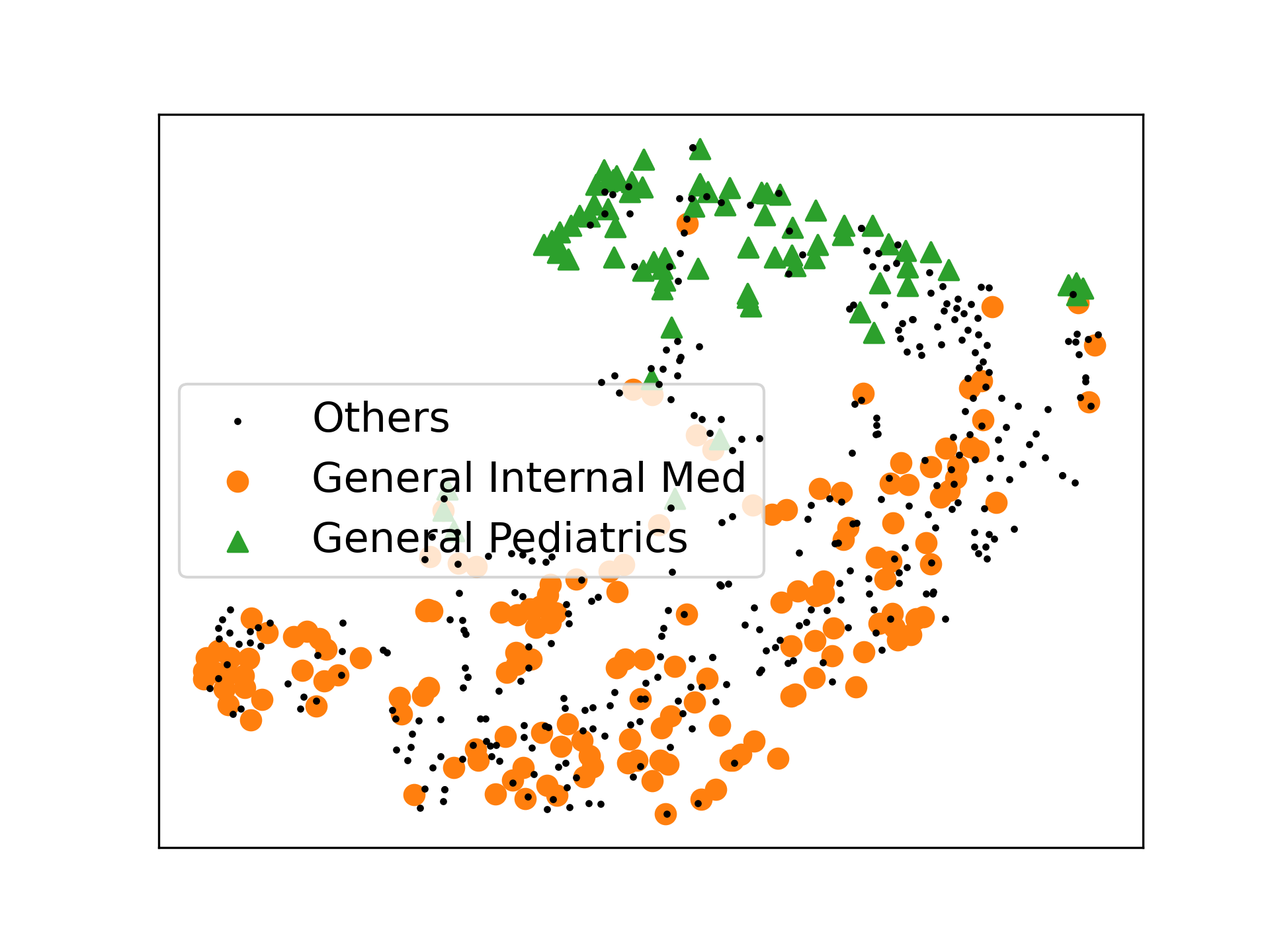}
\caption{Doctor embeddings learned by \ourmethod.
On average pediatricians are further away from the rest of the doctors compared to doctors in Internal Medicine.
}
\label{fig:DECEnt_doctor_2d}
\end{figure}

One of the findings from the pairwise dispersion also confirms what is generally known about hospital operations. 
For example, doctors in general internal medicine and anesthesia interact with a wide variety of patients, and thus doctors in these specialties are close to those in other specialties. On the other hand, patients of doctors in Family Practice and Otolaryngology (ears, nose, and throat) tend not to require referral to other specialties. 
As a result, doctors in these specialties are relatively far away from doctors in other specialties.
In Figure \ref{fig:DECEnt_doctor_2d}, we plot the doctor embeddings learned by DECEnt projecting them to 2-d using t-SNE \cite{van2008visualizing}.


\section{Related Work}
\label{section:related_works}

\par \noindent \textbf{Network Embeddings for Healthcare Analytics.}
Network embedding~\cite{grover2016node2vec,perozzi2014deepwalk} has gained much research interest lately. 
Recently, several approaches to learn embeddings of nodes in a dynamic network have been proposed. 
These approaches aim to capture both structural similarity and temporal evolution
\cite{Torricelli2020}. 


Similar approaches have been explored for medical data. 
A class of approaches
\cite{choi2018mime}
preserve the similarity of the medical codes in consecutive hospital visits by the same patient. 
eNRBM \cite{Tran2015eNRBM} uses restricted Boltzmann Machines, and \cite{Zhu2016CNN} uses convolutional neural network to represent abstract medical concepts. 

\par \noindent \textbf{Coevolving Networks.}
User-item interaction based embedding has gained a lot of attention recently. It has become a powerful tool for representing the evolution of users and items based on dynamic interactions
\cite{FarajtabarWGLZS15COEVOLVE}.
DeepCoevolve \cite{dai2016deep} uses RNN to learn the user and item embedding through 
the complex mutual influence in any interaction over the time. JODIE \cite{kumar2019predicting} extends \cite{dai2016deep} by adding a new projection operator that can predict user-item interaction at any future point of the time.

\par \noindent \textbf{Healthcare Analytics.} An area closely related to the current paper is that of Healthcare Analytics. Li et al. explored applicability of machine learning for CDI prediction using manual feature engineering~\cite{li2019using}.
Several other approaches have been proposed for case detection tasks 
\cite{makar2018learning}. 
A separate line of work focuses on mortality prediction~\cite{sherman2017leveraging}.
Other loosely related works include outbreak detection~\cite{adhikari2019fast}, missing infection inference~\cite{sundareisan2015hidden}, and architectural analysis~\cite{jang2019evaluating}.

\par \noindent \textbf{Autoencoders for Representation Learning.}
SDNE \cite{wang2016structural} uses autoencoders to preserve second-order proximity of the network.
NetRAs \cite{yu2018learning} uses graph encoder-decoder framework using LSTM networks and rooted random walks.
Deep Patient \cite{Miotto2016DeepPatient} uses stacked autoencoders to encode patients.

\section{Discussion and Limitations}
\label{section:discussion}
This paper proposes \ourmethod, a novel approach for learning heterogeneous dynamic embeddings of patients, doctors, rooms, and medications from diverse hospital operation data streams.
These embeddings capture similarities among entities 
based on static attributes and dynamic interactions. 
Consequently, these embeddings can serve as input to a variety of prediction tasks to improve clinical decision-making and patient care.
Our results show that on a variety of prediction tasks \ourmethod substantially outperforms baselines and produces embeddings meaningful to clinical experts. 

As we see it, our work has limitations.
All our patient-physician interactions are associated with procedures. 
While such procedure-related interactions are essential, we need to consider a more extensive, richer set of patient-physician interactions.
Such interactions can be extracted from the clinical notes from hospitals.
Each clinical note provides much context for each interaction, and one possible way to label the interaction with this context is to compute embeddings of clinical notes and attach these as weights or features to the interactions.
In the longer term, we are interested in deploying \ourmethod on top of the existing electronic medical record system at the hospital to provide clinical support. To reach this goal, we need a way to make the embeddings learned by \ourmethod interpretable to healthcare professionals. Specifically, we need to identify and explain the factors that cause two healthcare entities to be close to (or far from) each other. We propose to do so by introducing prototypes~\cite{dovsilovic2018explainable} and adding an explainability module for feature ablation.

Besides the future work mentioned above, 
we see a direction to expand our work. 
We can apply \ourmethod to other prediction tasks in the healthcare setting. 
For example, predicting the risk of readmission (see \cite{min2019predictive})
prior to discharge and predicting the 
length of stay in the hospital early during a patient visit (see \cite{TURGEMAN2017376}) are both tasks that can enable additional clinical resources for high-risk patients.

\par \noindent \textbf{Acknowledgements:}
The authors acknowledge funding from the CDC MInD Healthcare Network grant U01CK000594 and NSF grant 1955939.
The authors thank feedback from other University of Iowa CompEpi group members.

\newpage
\bibliographystyle{IEEEtran}

\end{document}